\documentclass{article}
\usepackage[preprint]{spconf,amsmath,graphicx}

\usepackage{times}
\usepackage{epsfig}
\usepackage{graphicx}
\usepackage{amsmath}
\usepackage{amssymb}
\usepackage{url}
\usepackage{hyperref}
\hypersetup{
    colorlinks=true,
    linkcolor=blue,
    filecolor=magenta,      
    urlcolor=magenta,
    pdftitle={Overleaf Example},
    pdfpagemode=FullScreen,
    }

\usepackage{framed,multirow}

\usepackage{amssymb}
\usepackage{latexsym}

\usepackage{amsmath}
\usepackage{makecell}
\usepackage{multirow}
\usepackage{booktabs}

\title{Coordinative Learning with Ordinal and Relational Priors for Volumetric Medical Image Segmentation}
\name{Haoyi Wang}
\address{School of Engineering, Computing, and Mathematics, University of Plymouth}
\begin{document}
%
\maketitle
\begin{abstract}
Volumetric medical image segmentation presents unique challenges due to the inherent anatomical structure and limited availability of annotations. While recent methods have shown promise by contrasting spatial relationships between slices, they rely on hard binary thresholds to define positive and negative samples, thereby discarding valuable continuous information about anatomical similarity. Moreover, these methods overlook the global directional consistency of anatomical progression, resulting in distorted feature spaces that fail to capture the canonical anatomical manifold shared across patients. To address these limitations, we propose Coordinative Ordinal-Relational Anatomical Learning (CORAL) to capture both local and global structure in volumetric images. First, CORAL employs a contrastive ranking objective to leverage continuous anatomical similarity, ensuring relational feature distances between slices are proportional to their anatomical position differences. In addition, CORAL incorporates an ordinal objective to enforce global directional consistency, aligning the learned feature distribution with the canonical anatomical progression across patients. Learning these inter-slice relationships produces anatomically informed representations that benefit the downstream segmentation task. Through this coordinative learning framework, CORAL achieves state-of-the-art performance on benchmark datasets under limited-annotation settings while learning representations with meaningful anatomical structure. Code is available at \url{https://github.com/haoyiwang25/CORAL}.

\end{abstract}
\section{Introduction}
\label{sec:intro}



Volumetric medical image segmentation is a core task in medical image analysis, enabling automated diagnosis, treatment planning, and disease monitoring~\cite{litjens2017survey}. However, obtaining annotations is particularly challenging due to the labor-intensive nature of slice-by-slice annotation and the requirement for highly specialized domain expertise~\cite{tajbakhsh2020embracing}. Recently, semi-supervised methods leveraging contrastive pre-training have emerged as a promising paradigm to address this data scarcity challenge~\cite{chen2020simple}. However, general-purpose contrastive learning methods designed for natural images often yield suboptimal results when directly applied to medical imaging~\cite{chaitanya2020contrastive}.

The primary issue lies in the false negative problem, where slices containing the same anatomical structure from different patients are incorrectly treated as negative pairs and pushed apart in the feature space~\cite{chaitanya2020contrastive}. To alleviate this problem, Positional Contrastive Learning (PCL)~\cite{zeng2021positional} leverages the intrinsic spatial organization of volumetric medical images, based on the fact that adjacent slices in a 3D volume typically contain similar anatomical structures, while distant slices exhibit dissimilar content.


Despite its success, PCL has a fundamental limitation due to its reliance on a hard binary threshold, which creates an abrupt categorical boundary in the feature space. Moreover, this threshold is highly dataset-dependent, requiring manual tuning for each imaging modality or anatomical region. Self-supervised Alignment Learning (SAL)~\cite{li2024self}, the most recently proposed method, also follows this same design paradigm.


Beyond this threshold-based limitation, existing methods overlook a fundamental structural property of volumetric medical images, limiting their ability to capture this global anatomical structure. Specifically, the anatomical progression along an organ follows a consistent directional trajectory across different patients. For example, cardiac MRI scans progress from the apex to the base of the heart in a predictable manner, regardless of individual anatomical variations. This property suggests that slice embeddings should not only preserve local pairwise similarities but also exhibit globally aligned directional patterns that reflect the canonical anatomical manifold shared across patients. 


In this paper, we propose Coordinative Ordinal-Relational Anatomical Learning (CORAL) to address both limitations. CORAL consists of two complementary components: a Relational Anatomical Learning (RAL) objective and an Ordinal Anatomical Learning (OAL) objective. To address the hard threshold problem, RAL replaces the binary thresholds in PCL and SAL with a relational ranking loss that enforces a continuous relationship where slices closer in anatomical position exhibit greater similarity in the learned embedding space, eliminating the need for manual threshold tuning. To address the lack of global directional consistency, OAL introduces an ordinal loss that enforces consistent anatomical trajectories across patients, ensuring that embeddings not only preserve local relational orderings but also align along a shared canonical anatomical direction. By jointly optimizing these two complementary objectives, CORAL preserves continuous anatomical relationships at the local level while maintaining consistent directional alignment at the global level. 

While segmentation ultimately operates on individual 2D slices, learning inter-slice 3D relationships during pre-training produces anatomically informed representations that transfer effectively to the downstream task. This knowledge acts as an structural prior that enhances segmentation performance, particularly in limited-annotation settings.

Our key contributions are:
\begin{itemize}
\item We propose Coordinative Ordinal-Relational Anatomical Learning (CORAL) to learn anatomically meaningful representations in volumetric medical images.
\item We introduce a Relational Anatomical Learning (RAL) objective that enforces continuous similarity relationships between slices based on their anatomical proximity, eliminating the need for manual threshold tuning.
\item We propose an Ordinal Anatomical Learning (OAL) objective that enforces consistent anatomical trajectories aligned across patients, capturing the canonical directional progression inherent in volumetric data.
\item We demonstrate state-of-the-art performance on multiple segmentation benchmark datasets, with substantial improvements in limited-annotation settings.
\end{itemize}

The rest of this paper is organized as follows. In Section~\ref{sec:method}, we present the technical details of CORAL. In Section~\ref{sec:experiments}, we describe experimental settings and results. Finally, we conclude in Section~\ref{sec:conclusion}.

\section{Coordinative Ordinal-Relational Anatomical Learning}
\label{sec:method}

In this section, we provide a detailed explanation of CORAL by first formulating the anatomical representation learning for volumetric medical imaging followed by reviewing PCL. Then, we progressively introduce RAL, OAL, and finally CORAL.

\subsection{Problem Formulation}

Given a dataset of $X$ volumetric medical images $\{V^{x}\}_{x=1}^{X}$, where each volume $V^{x}$ consists of $Z^{x}$ axial slices, we follow a two-stage semi-supervised learning pipeline. The first stage performs pre-training on unlabeled volumes to learn an encoder $f: \mathbb{R}^{H \times W} \rightarrow \mathbb{R}^d$ that maps 2D slices of dimension $H \times W$ to a $d$-dimensional feature space. In the second stage, the trained encoder is used to fine-tune a segmentation network on a small set of labeled volumes with supervised loss to produce pixel-wise segmentation masks. For volume $V^{x}$, we denote the slice at axial index $z$ as $s_z^{x} \in \mathbb{R}^{H \times W}$ and assign it a normalized position $p_z = z / Z^{x} \in [0,1]$. This normalization enables cross-volume comparison despite varying numbers of slices across different volumes.

The core challenge in the pre-training stage is to encode two complementary structural properties: (1) \textit{local relational similarity}, where anatomically proximate slices should have similar embeddings, and (2) \textit{global ordinal consistency}, where the anatomical progression should follow aligned trajectories in the embedding space across patients. Formally, we seek to learn $f(\cdot)$ such that for any two slices with learned embeddings $h_i^{x} = f(s_i^{x})$ and $h_j^{y} = f(s_j^{y})$, from the same or different volumes, their cosine similarity $\text{sim}(h_i, h_j)$ satisfies:
\begin{equation}
\text{sim}(h_i^{x}, h_j^{y}) \propto 1 / |p_i - p_j|,
\end{equation}
while the embedding space maintains consistent directional alignment with the anatomical progression across the entire dataset.

It is important to note that while segmentation is fundamentally an intra-slice task operating on individual 2D slices, learning inter-slice relationships during pre-training provides crucial anatomical context that enhances the encoder's representational capacity~\cite{zeng2021positional, zhou2019models}. This anatomical awareness enables the learned features to better generalize to the downstream segmentation task even when fine-tuning with limited labeled data~\cite{chen2020big}. 

For notational simplicity, we henceforth omit the volume superscript when the context is clear.

\subsection{Review of Positional Contrastive Learning}

PCL leverages the spatial organization of volumetric scans by defining positive and negative pairs based on their positional differences. Specifically, two slices $s_i$ and $s_j$ form a positive pair if $|p_i - p_j| < \delta$, where $\delta$ is a dataset-specific threshold hyperparameter; otherwise, they form a negative pair. PCL optimizes the following contrastive loss:
\begin{equation}
\mathcal{L}_{PCL} = -\sum_{i} \frac{1}{|\mathcal{A}_i|} \sum_{a \in \mathcal{A}_i} \log \frac{\exp(\text{sim}(h_i, h_a)/\tau)}{\sum_{k \neq i} \exp(\text{sim}(h_i, h_k)/\tau)},
\end{equation}
where $\mathcal{A}_i$ denotes the set of positive sample indices for anchor $i$, and $\tau$ is the temperature hyperparameter.

Despite its effectiveness, PCL suffers from two fundamental limitations. First, the binary threshold $\delta$ creates a hard boundary between positive and negative pairs, discarding continuous information about the degree of anatomical similarity. Two slice pairs with positional differences of $\delta - \epsilon$ and $\delta + \epsilon$ (for arbitrarily small $\epsilon > 0$) are treated categorically differently despite negligible spatial separation. Moreover, the optimal value of $\delta$ must be manually tuned for each dataset, as it varies significantly across imaging modalities and anatomical regions. Second, PCL only enforces local pairwise relationships without imposing global ordinal consistency across patients. While it ensures nearby slices have similar embeddings, it does not guarantee that anatomical trajectories align in a canonical direction across the dataset, potentially limiting the model's ability to capture the shared anatomical manifold across patients.

\subsection{Relational Anatomical Learning}

We introduce RAL to directly address the first limitation of PCL. RAL replaces PCL's binary classification of slice pairs with a continuous listwise ranking mechanism. Specifically, we adapt the ranking framework from Rank-N-Contrast (RNC)~\cite{zha2023rank}. However, while RNC ranks samples based on label distances in classification tasks, RAL ranks slices based on their anatomical position differences in volumetric medical images, creating a model that is sensitive to the fine-grained continuous relationship between spatial proximity and feature similarity.

For an anchor slice $s_i$ and any other slice $s_j$ in a batch, we form a positive pair $(s_i,s_j)$. The negative set $\mathcal{N}_r(s_i,s_j)$ for this pair consists of all slices $s_k$ whose positional distance from $s_i$ is greater than or equal to that of $s_j$:
\begin{equation}
\mathcal{N}_r(s_i,s_j) = \{s_k \mid k \ne i, |p_i - p_k| \ge |p_i - p_j|\}.
\end{equation}

The relational loss in RAL encourages the embedding $h_j$ to be more similar to the anchor $h_i$ than any negative sample $h_k = f(s_k)$ from the set $\mathcal{N}_r(s_i, s_j)$:
\begin{equation}
\mathcal{L}_{RAL} = -\sum_{i=1}^{N} \sum_{\substack{j=1 \\ j \neq i}}^{N} \log \frac{\exp(\text{sim}(h_i, h_j)/\tau)}{\sum_{s_k \in \mathcal{N}_r(s_i, s_j)} \exp(\text{sim}(h_i, h_k)/\tau)}.
\end{equation}

This listwise ranking loss enforces a continuous relational ordering: if position $p_j$ is closer to $p_i$ than $p_k$, then the embedding $h_j$ must be more similar to $h_i$ than $h_k$. This directly captures the continuous nature of anatomical variation, eliminating the need for the arbitrary and dataset-dependent threshold hyperparameter $\delta$. The resulting feature space exhibits a substantially finer-grained and more geometrically accurate structure.

\subsection{Ordinal Anatomical Learning}

While the relational prior ensures a fine-grained local ordering of the feature space, it does not impose any global structural consistency. To address this, we introduce an ordinal prior that forces the model to learn a single, globally aligned anatomical trajectory across all subjects. 

The core of this objective is a canonical direction vector, $c_{anat} \in \mathbb{R}^d$, which represents the overall anatomical progression from one end of a volume to the other. $c_{anat}$ is initialized randomly and learned during the pre-training stage.

To enforce global directional consistency in the feature space, previous approaches~\cite{wang2025age} partition the continuous data distribution into discrete groups and use group representatives, e.g., cluster centers, to guide the learning of the global direction vector. However, defining such discrete groupings for volumetric medical images lacks both practical feasibility and theoretical justification, as anatomical progression is inherently continuous. Therefore, we formulate the OAL objective as a direct alignment loss based on negative cosine similarity. Specifically, for any pair of slice embeddings $(h_i, h_j)$ where $p_j > p_i$, we define the unit direction vector as $v^{(i,j)} = (h_j - h_i) / ||h_j - h_i||_2$. The OAL objective maximizes the cosine similarity between each pairwise direction vector $v^{(i,j)}$ and the learnable global canonical direction vector $c_{\text{anat}}$:
\begin{equation}
\mathcal{L}_{OAL} = - \mathbb{E}_{i,j \mid p_j > p_i} \left[ \text{sim}(v^{(i,j)}, c_{\text{anat}}) \right].
\end{equation}

By minimizing this loss, all pairwise direction vectors $v^{(i,j)}$ are encouraged to align with the canonical direction vector $c_{\text{anat}}$, which in turn learns a consistent and globally aligned embedding trajectory for anatomical progression across all patient volumes.

\subsection{Coordinative Ordinal-Relational Anatomical Learning}

CORAL combines RAL and OAL to learn a well-structured anatomical manifold. The relational loss $\mathcal{L}_{RAL}$ ensures fine-grained local similarity relationships, while the ordinal loss $\mathcal{L}_{OAL}$ enforces global directional consistency and alignment across patients. The complete CORAL loss function is:
\begin{equation}
\mathcal{L}_{CORAL} = \mathcal{L}_{RAL} + \mathcal{L}_{OAL}.
\end{equation}
We do not apply weighting factors to either component, as both local relational structure and global directional alignment are equally important for learning anatomically meaningful representations.

\section{Experiments}
\label{sec:experiments}

\subsection{Datasets}

We evaluate our method on two benchmark volumetric medical image segmentation datasets: ACDC (Automated Cardiac Diagnosis Challenge)~\cite{bernard2018deep} and CHD (Congenital Heart Disease)~\cite{xu2019whole}. We strictly follow the training strategy in~\cite{zeng2021positional}. Specifically, in the pre-training stage, we pre-train a U-Net encoder~\cite{ronneberger2015u} on the entire dataset without using any labels. After pre-training, the encoder is frozen, and the U-Net decoder is fine-tuned using a small number of labeled samples. Fine-tuning is performed using 5-fold cross-validation. For each fold, a specific number of patients, denoted as M, are randomly sampled to serve as the labeled training set, and the model is evaluated on the remaining patients.


\subsection{Implementation Details}

For both datasets, we normalize the intensity of each 3D volume and perform two standardization steps for every 2D slice. In specific, the ACDC dataset slices are resampled to a resolution of 1.25×1.25mm and padded to a size of 352×352 pixels, while the CHD dataset slices are resampled to 1.0×1.0mm and padded to 512×512 pixels. As the images in both datasets were already roughly aligned, no further alignment techniques were applied.

During the pre-training stage, the model was trained for 300 epochs with a batch size of 32. The training utilized an SGD optimizer with a learning rate of 0.1 and a cosine learning rate scheduler. $\tau$ was set to 0.1. In the fine-tuning stage, the segmentation network was trained for 100 epochs using a standard cross-entropy loss with a batch size of 5. The Adam optimizer was used with a cosine scheduler and an initial learning rate of 5×10$^{-5}$.

\begin{table}
\begin{center}
\label{tab:acdc}
\caption{Comparison of the CORAL with baseline models on the ACDC dataset. The best results are highlighted in bold.}
\begin{tabular}{p{0.125\textwidth}p{0.09\textwidth}p{0.09\textwidth}p{0.09\textwidth}}
\toprule
\hfil{Method} & \hfil{M=2} & \hfil{M=6} & \hfil{M=10} \\ \midrule
\hfil{GCL \cite{chaitanya2020contrastive}} & \hfil{0.636(.05)} & \hfil{0.803(.04)} & \hfil{0.872(.01)} \\
\hfil{CAiD \cite{taher2022caid}} & \hfil{0.483(.11)} & \hfil{0.822(.02)} & \hfil{0.879(.02)} \\
\hfil{KAF \cite{yang2023keypoint}} & \hfil{0.741(.03)} & \hfil{0.873(.01)} & \hfil{0.895(.01)} \\
\hfil{SAL \cite{li2024self}} & \hfil{0.723(.04)} & \hfil{0.869(.02)} & \hfil{0.890(.01)} \\
\hfil{PCL \cite{zeng2021positional}} & \hfil{0.671(.06)} & \hfil{0.850(.01)} & \hfil{0.885(.01)} \\
\midrule
\hfil{w/ $\mathcal{L}_{RAL}$} & \hfil{0.718(.04)} & \hfil{0.863(.02)} & \hfil{0.860(.01)} \\
\hfil{w/ $\mathcal{L}_{OAL}$} & \hfil{0.700(.06)} & \hfil{0.824(.03)} & \hfil{0.837(.02)} \\
\hfil{CORAL (Ours)} & \hfil{\textbf{0.746(.05)}} & \hfil{\textbf{0.886(.02)}} & \hfil{\textbf{0.903(.02)}} \\
\bottomrule
\end{tabular}
\end{center}
\end{table}

\begin{table}
\begin{center}
\label{tab:chd}
\caption{Comparison of the CORAL with baseline models on the CHD dataset. The best results are highlighted in bold.}
\begin{tabular}{p{0.125\textwidth}p{0.09\textwidth}p{0.09\textwidth}p{0.09\textwidth}}
\toprule
\hfil{Method} & \hfil{M=2} & \hfil{M=6} & \hfil{M=10} \\ \midrule
\hfil{GCL \cite{chaitanya2020contrastive}} & \hfil{0.255(.10)} & \hfil{0.564(.04)} & \hfil{0.646(.03)} \\
\hfil{CAiD \cite{taher2022caid}} & \hfil{0.265(.08)} & \hfil{0.581(.06)} & \hfil{0.647(.04)} \\
\hfil{KAF \cite{yang2023keypoint}} & \hfil{0.392(.06)} & \hfil{0.636(.06)} & \hfil{\textbf{0.693(.03)}} \\
\hfil{SAL \cite{li2024self}} & \hfil{0.348(.04)} & \hfil{0.572(.06)} & \hfil{-} \\
\hfil{PCL \cite{zeng2021positional}} & \hfil{0.356(.08)} & \hfil{0.600(.06)} & \hfil{0.661(.05)} \\
\midrule
\hfil{w/ $\mathcal{L}_{RAL}$} & \hfil{0.388(.08)} & \hfil{0.632(.05)} & \hfil{0.681(.05)} \\
\hfil{w/ $\mathcal{L}_{OAL}$} & \hfil{0.365(.10)} & \hfil{0.604(.06)} & \hfil{0.649(.07)} \\
\hfil{CORAL (Ours)} & \hfil{\textbf{0.400(.08)}} & \hfil{\textbf{0.657(.06)}} & \hfil{0.684(.05)} \\
\bottomrule
\end{tabular}
\end{center}
\end{table}

\subsection{Comparison with Baseline Methods}




To demonstrate its effectiveness, we compare CORAL against recent state-of-the-art semi-supervised methods for volumetric medical image segmentation. As shown in Tables 1 and 2, CORAL consistently outperforms all baselines across most settings. The only exception occurs on the CHD dataset when M = 10. We attribute this to CORAL's exclusive focus on inter-slice relationships, whereas Keypoint-Augmented Fusion (KAF)~\cite{yang2023keypoint} additionally models intra-slice spatial correspondence through keypoint matching, which may provide complementary information for this particular setting.

Examining the ablation variants, we observe that using only $\mathcal{L}_{RAL}$ provides substantial gains over PCL by eliminating the binary threshold, validating our hypothesis about the importance of continuous relational modeling. Using only $\mathcal{L}_{OAL}$ shows more modest improvements, suggesting that global ordinal alignment requires fine-grained local relational structure to be fully effective. CORAL, which combines both components, achieves the best performance in all settings, demonstrating the complementary benefits of local relational ranking and global ordinal constraints.





\begin{figure}[!t]
\centering
\includegraphics[width=3.2in]{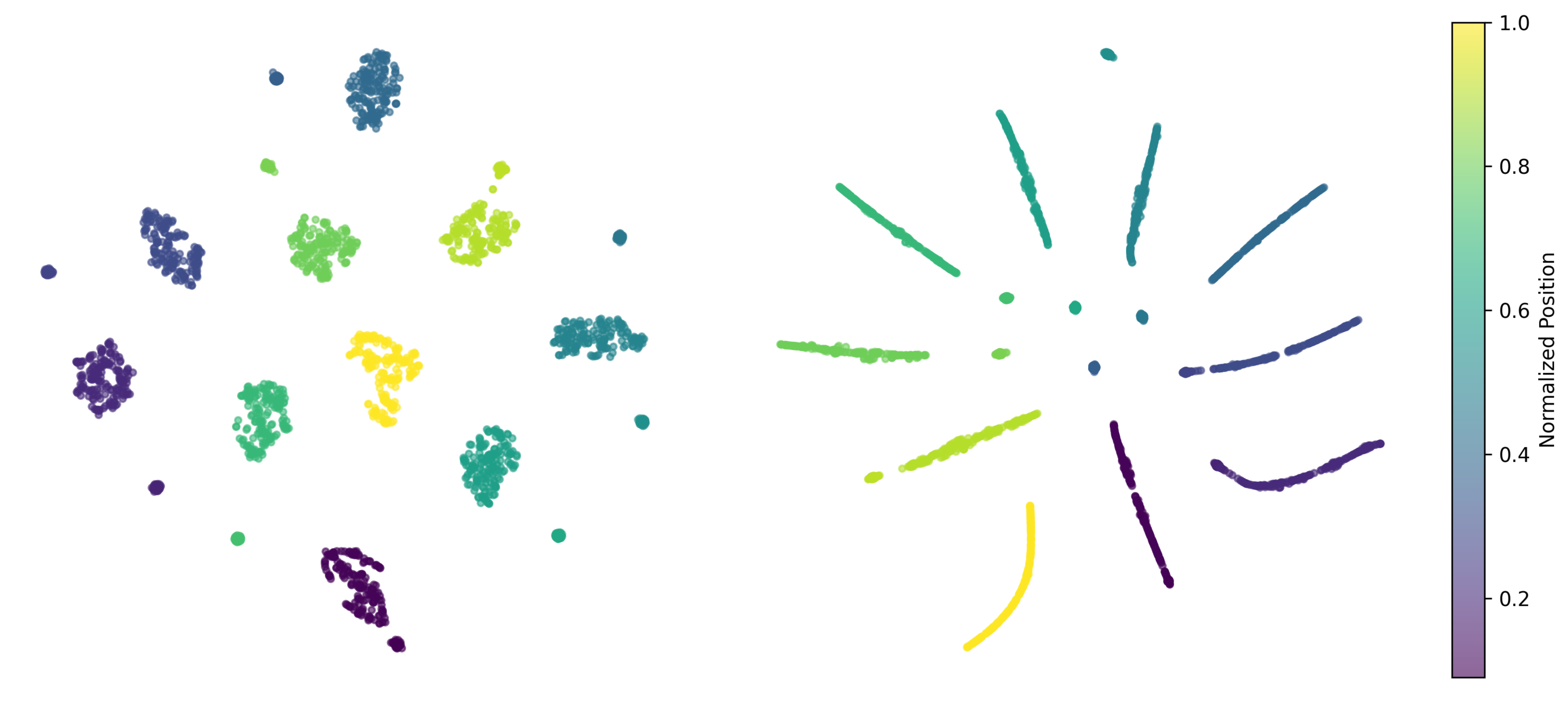}
\caption{Comparison of learned feature space from PCL (left) and CORAL (right) on the ACDC dataset.}
\label{fig:compare}
\end{figure}

Fig. 1 offers a visualization of the feature spaces learned by PCL and our proposed CORAL framework. The t-SNE plot for PCL shows a fragmented and disordered representation. While some local color gradients are visible within these clusters, there is no overarching global structure or alignment between them. In contrast, CORAL learns smooth, continuous, and well-ordered trajectories. The color gradient progresses consistently along these paths, providing clear visual evidence that CORAL successfully learns a feature space where movement along a trajectory directly corresponds to the anatomical progression of slices. 

\section{Conclusion}
\label{sec:conclusion}

In this paper, we propose Coordinative Ordinal-Relational Anatomical Learning (CORAL), a novel semi-supervised framework for volumetric medical image segmentation. By combining relational ranking with ordinal directional constraints, CORAL preserves both fine-grained local similarity and global trajectory alignment. Experiments on benchmark datasets demonstrate that CORAL achieves state-of-the-art performance in limited-annotation settings while learning anatomically meaningful representations.



\bibliographystyle{IEEEbib}
\bibliography{strings,refs}

\end{document}